\documentclass{article}
\usepackage{mlspconf,amsmath,graphicx}
\usepackage{amsmath,graphicx}
\usepackage{dsfont}
\usepackage{float}
\usepackage{xfrac}
\usepackage{color}
\usepackage{psfrag}
\usepackage{amsthm}
\usepackage{subfigure}
\usepackage[lined,boxed,commentsnumbered]{algorithm2e}
\usepackage{enumerate}
\usepackage{import}
\usepackage{epsfig}
\usepackage{amssymb}
\usepackage{pgfplots}
\usepackage{graphicx}
\usepackage{sistyle}
\usepackage{url}

\DeclareMathAlphabet{\bm}{OML}{cmr}{bx}{it}
\DeclareMathAlphabet{\mathsf}{OT1}{cmss}{m}{n}
\DeclareMathAlphabet{\bs}{T1}{cmss}{bx}{sl}
\DeclareMathAlphabet{\ms}{T1}{cmss}{m}{sl}
\DeclareMathAlphabet{\mathpzc}{OML}{zplm}{m}{it}

\newcommand{\bg}[1]{\boldsymbol #1} 

\title{Physics-informed neural networks for pathloss prediction}
\name{Steffen Limmer, Alberto Martinez Alba, Nicola Michailow \thanks{This work was supported by the German Federal Ministry of Education and Research (BMBF) under grant 16KIS1103.}}
\address{Siemens AG, Technology, Otto-Hahn-Ring 6, 81739 Munich, Germany}

\begin{document}
%
\maketitle
\begin{abstract}
This paper introduces a physics-informed machine learning approach for pathloss prediction. This is achieved by including in the training phase simultaneously (i) physical dependencies between spatial loss field and (ii) measured pathloss values in the field. It is shown that the solution to a proposed learning problem improves generalization and prediction quality with a small number of neural network layers and parameters. The latter leads to fast inference times which are favorable for downstream tasks such as localization. Moreover, the physics-informed formulation allows training and prediction with a small amount of training data which makes it appealing for a wide range of practical pathloss prediction scenarios.
\end{abstract}
\section{Introduction}
\label{sec:intro}

Learning and prediction of spatial loss fields and pathloss maps in radio technologies is an important problem used for many downstream tasks such as localization and quality-of-service prediction (data rate, latency, upcoming connection loss, etc.). These tasks are often based on received signal strength indicator (rssi) of the signal. For example, rssi is used to trigger handovers and to enable basestation association for load balancing purposes. Thus, utilizing rssi information for learning and prediction of pathloss maps is attractive since pathloss measurements are already available in many wireless protocols. One drawback of rssi is a high measurement and calibration effort which is necessary in order to achieve a satisfactory accuracy for pathloss prediction or solving downstream tasks such as localization to high accuracy.

On the other hand, model-based simulation methods have been used to simulate wave propagation effects in structured environments as typically found inside buildings and dense building areas. Such propagation environments are typically composed of individual objects such as walls, furniture, machinery or building materials for which physical transmission properties such as transmission coefficients at various frequencies are available from measurement campaigns \cite{Anderson2004}. However, using model-based simulations for solving downstream tasks such as localization is typically difficult due to model mismatch and lacking support for automatically calculated gradients.

\begin{figure}[t!]
	\includegraphics[width=\columnwidth]{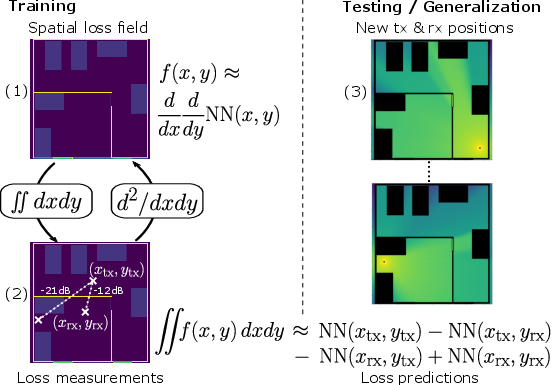}
	\caption{Physics-informed pathloss prediction. (1) The spatial loss field is approximated by the (mixed) derivative of an NN. (2) Pathloss measurements are approximated by signed sums of NN evaluations. (3) At test time, loss predictions are made for arbitrary tx, rx positions by signed sums of NN evaluations.}
	\label{fig:teaser}
    \vspace{-1em}
\end{figure}

This paper proposes a method to learn spatial loss fields and perform low-complexity physics-informed prediction based on modelling of site-information and recorded pathloss measurements. The proposed method is illustrated in Fig. \ref{fig:teaser} and uses a physics-based problem formulation to achieve better generalization to new transmit and receive locations and enables physically consistent solutions of downstream tasks such as localization.

\section{Prior Art}
\subsection{Model-based simulation}
Model-based simulations such as WinProp\footnote{https://www.altair.com/feko-applications} or Pylayers \cite{Amiot2013} are often based on raytracing or on finite-difference time domain techniques and can be used to model spatial loss fields and simulate pathloss maps. Simulations are typically very costly with reported runtimes of approximately \SI{1}{s} for simulations based on the dominant path model, and \SI{10}{s} for simulations based on intelligent ray tracing \cite{Levie2021}. Due to the long runtime, it is difficult to solve downstream tasks such as localization with model-based simulations, e.g., robot navigation or robot collaboration in highly dynamic environments like factories. In addition, simulators usually lack support for automatic differentiation which additionally slows down downstream optimization tasks. Simulations need to be calibrated to specific measurement hardware and need fine-grained modelling of the site which is costly. In addition, there is typically only limited or no support to model partial information about the site geometry.

\subsection{Machine learning approaches}
Machine learning approaches to pathloss prediction and localization can be based, e.g., on convolutional networks \cite{Levie2021}. The work \cite{Levie2021} proposes a UNet architecture trained to mimic the rssi prediction of a model-based simulator which reduces runtime to approximately \SI{1}{ms} and supports automatic differentiation for downstream tasks such as localization. On the other hand, it is not directly possible to include physical propagation constraints and real-world measurements. As a standard convolutional neural network is not aware of underlying physical dependencies, the network requires a high number of trainable parameters (exceeding 1 million) which results in long training times. In addition, the unawareness of physical dependencies may result in artifacts and inconsistencies in the reconstructed pathloss maps.

\section{Physical propagation model}
\label{sec:phys}
In the following we introduce a physics-informed learning problem, where a single neural network is trained jointly on pathloss (rssi) measurements and information about the underlying spatial loss field. Following \cite{Patwari2008}, the rssi in dB between a transmitter located at $\bm{x}_\text{tx}=[x_\text{tx},y_\text{tx}]$ and a receiver located at $\bm{x}_\text{rx}=[x_\text{rx},y_\text{rx}]$ is measured in dB modelled by 
\begin{equation}\label{equ:pl0}
\text{RSSI}(\bm{x}_\text{tx},\bm{x}_\text{rx}) = G_0 - \gamma \log_{10}\lVert \bm{x}_\text{tx} - \bm{x}_\text{rx} \rVert - \text{ISLF}(\bm{x}_\text{tx},\bm{x}_\text{rx}).
\end{equation}
In \eqref{equ:pl0}, the first component $G_0$ denotes the transmit power, the second component models the free-space component of the pathloss and the third component is the integrated spatial loss field $\text{ISLF}$. The $\text{ISLF}$ models additional attenuation by obstacles as a (weighted) integral over a spatial loss field $\text{SLF}$ according to
\begin{equation}\label{equ:pl1}
\text{ISLF}(\bm{x}_\text{tx},\bm{x}_\text{rx}) = \int_{\bm{x}_\text{rx}}^{\bm{x}_\text{tx}} w(\bm{x}) \text{SLF}(\bm{x}) \ d\bm{x}.
\end{equation}
In this work, we set the weight function $w(\bm{x})$ as in computer tomography applications to a Dirac delta distribution supported on the line through $\bm{x}_\text{tx}$ and $\bm{x}_\text{rx}$. 

The transmit power $G_0$ and free-space component parameter $\gamma$ can be assumed known as they can be measured or estimated using conventional approaches. Hence, the main problem is to obtain a tractable learning approach involving SLF and ISLF. The latter explicitly encodes the transmission of radio waves travelling between different transmit-receive pairs $(\bm{x}_\text{rx},\bm{x}_\text{tx})$ through common building materials.

ISLF training data is obtained from rssi measurements between different tx-rx pairs according to
\begin{align*}
\text{ISLF}(\bm{x}_\text{tx},\bm{x}_\text{rx}) = G_0 - \text{RSSI}(\bm{x}_\text{tx},\bm{x}_\text{rx}) - \gamma \log_{10}\lVert \bm{x}_\text{tx} - \bm{x}_\text{rx} \rVert
\end{align*}
resulting in a dataset
\begin{equation}
\mathcal{D}_\text{ISLF}:= (\bm{x}^{(i)}_\text{tx},\bm{x}^{(i)}_\text{rx},\text{ISLF}(\bm{x}^{(i)}_\text{tx},\bm{x}^{(i)}_\text{rx}))_{i \in \mathcal{I}},
\end{equation}
where $(\cdot)^{(i)}$ denotes the $i$-th element in the dataset.

SLF training data can be derived from site geometry information via transmission coefficients of common building materials such as drywall, whiteboard or glass as available in the literature (see Table \ref{tab:office_attenuation}, Fig. \ref{fig:SLF1}) resulting in a dataset
\begin{equation}
\mathcal{D}_\text{SLF}:= (\bm{x}^{(j)},\text{SLF}(\bm{x}^{(j)}))_{j \in \mathcal{J}}.
\end{equation}

\begin{table}[t!]
\begin{tabular}{|c|c|c|}
\hline 
Material &  Loss at 2.5 GHz & Loss at 60 GHz \\ 
\hline 
Drywall &  2.1 (dB/cm) & 2.4 (dB/cm)\\ 
\hline 
Whiteboard &  0.3 (dB/cm) & 5.0 (dB/cm) \\ 
\hline 
Glass &  20.0 (dB/cm) & 11.3 (dB/cm) \\ 
\hline 
\end{tabular}
\caption{Office material attenuation at 2.5 GHz and 60 GHz. (reproduced from \cite{Anderson2004})}
\label{tab:office_attenuation}
\end{table}

\section{Physics-informed learning of integral equations}
\subsection{Solving integral equations with neural networks}
Physics-informed neural networks (pinns) have been introduced recently in \cite{Raissi2019} to include physical constraints already in the design and training of neural networks. In the literature, pinns have been applied to partial-differential equations (pdes) such as the Navier–Stokes equations \cite{Raissi2019}, but pathloss prediction requires solving an integral equation \eqref{equ:pl0}. An approach to approximate integral equations via Monte-Carlo integration based on several model evaluations is used in \cite{Hennigh2021}. An approach for solving integral equations exactly has been presented recently in \cite{Mueller2020} but is based on a rather complex normalizing flow. This section outlines a simpler approach for solving exactly an integral equation based on standard neural networks and antiderivative rules for indefinite integration as explained, e.g., in \cite{Lindell2021}.

Conventional training of a neural network $\text{NN}(\bg{\theta},x,y)$ with parameters $\bg{\theta}$ for a bivariate function $f(x,y)$ is done by minimizing
\begin{equation}
\text{Loss}(\bg{\theta}) = \sum_i \phi\Bigl( f(x^{(i)},y^{(i)}) - \text{NN}(\bg{\theta},x^{(i)},y^{(i)}) \Bigr),
\end{equation}
where $\phi(\cdot)$ is an appropriate loss function such as squared, absolute or huber loss.
In order to solve an integral equation, the neural network $\text{NN}(\bg{\theta},x,y)$ is trained to calculate integrals 
\begin{equation}
\int_{x_\text{tx}}^{x_\text{rx}} \int_{y_\text{tx}}^{y_\text{rx}} f(x,y) \ dx dy
\end{equation} 
by minimizing the error
\begin{equation}
\text{Loss}(\bg{\theta}) = \sum_i \phi\Bigl( f(x^{(i)},y^{(i)}) - \frac{d}{dx} \frac{d}{dy} \text{NN}(\bg{\theta},x^{(i)},y^{(i)}) \Bigr),
\end{equation}
where the derivative $\frac{d}{dx} \frac{d}{dy} \text{NN}(\bg{\theta},x,y)$ can be calculated at training time and evaluated for different points $(x^{(i)},y^{(i)})$ in the training set using standard automatic differentiation. After optimizing the neural network parameters $\bg{\theta}^\star$, integrals over the rectangular domain $[x_\text{tx},x_\text{rx}]\times [y_\text{tx},y_\text{rx}]$ can be calculated according to
\begin{align}
& \int_{x_\text{tx}}^{x_\text{rx}} \int_{y_\text{tx}}^{y_\text{rx}} f(x,y) \ dx dy = \\
& \text{NN}(\bg{\theta},x_\text{tx},y_\text{tx}) - \text{NN}(\bg{\theta},x_\text{tx},y_\text{rx}) \nonumber \\
 - & \text{NN}(\bg{\theta},x_\text{rx},y_\text{tx}) + \text{NN}(\bg{\theta},x_\text{rx},y_\text{rx}). \nonumber
\end{align}

\subsection{Problem Formulation}
The joint physics-aware learning problem using SLF and ISLF training data can be posed via a novel training loss function. 

The general formulation assumes that the ISLF corresponds to a two-dimensional integral over the rectangular domain defined by $\bm{x}_\text{tx}:=(x_\text{tx},y_\text{tx})$ and $\bm{x}_\text{rx}:=(x_\text{rx},y_\text{rx})$, i.e.,
\begin{align}
\text{ISLF}(\bm{x}_\text{tx},\bm{x}_\text{rx}) = \int_{x_\text{tx}}^{x_\text{rx}} \int_{y_\text{tx}}^{y_\text{rx}} w(x,y) \text{SLF}(x,y) \ dxdy,
\end{align}
where $w(x,y)$ is a given weight function corresponding to different physical propagation models. Typical propagation models from the literature include the network shadowing, normalized ellipse, inverse area elliptical model \cite{Patwari2008,Hamilton2013} or the line integral model as in traditional computer tomography applications. 

In this work, we employ the line integral model so that the dimensionality of the integration can be reduced by the Radon transform \cite{Maier2018} according to
\begin{align*}
& \int_{x_\text{tx}}^{x_\text{rx}} \int_{y_\text{tx}}^{y_\text{rx}} \delta\Bigl( \overline{(x_\text{tx},y_\text{tx}),(x_\text{rx},y_\text{rx})} \Bigr) \text{SLF}(x,y) \ dxdy \\
 \equiv  & \int_{z_0}^{z_1} \text{SLF}(z,\alpha,s) \ dz,
\end{align*}
with $\delta\Bigl( \overline{(x_\text{tx},y_\text{tx}),(x_\text{rx},y_\text{rx})} \Bigr)$ denoting the Dirac distribution supported on the line through transmitter and receiver coordinates.

\begin{figure}
\centering \subfigure[]{
  \includegraphics[scale=0.14]{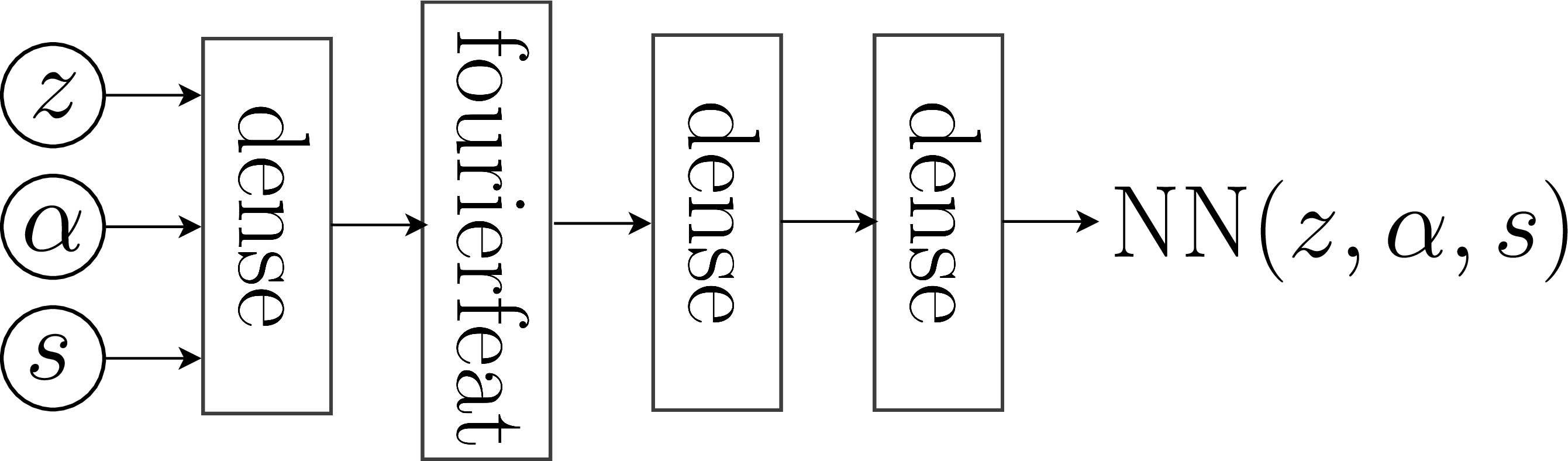}
} \hspace{-.55cm} \subfigure[]{
  \includegraphics[scale=0.14]{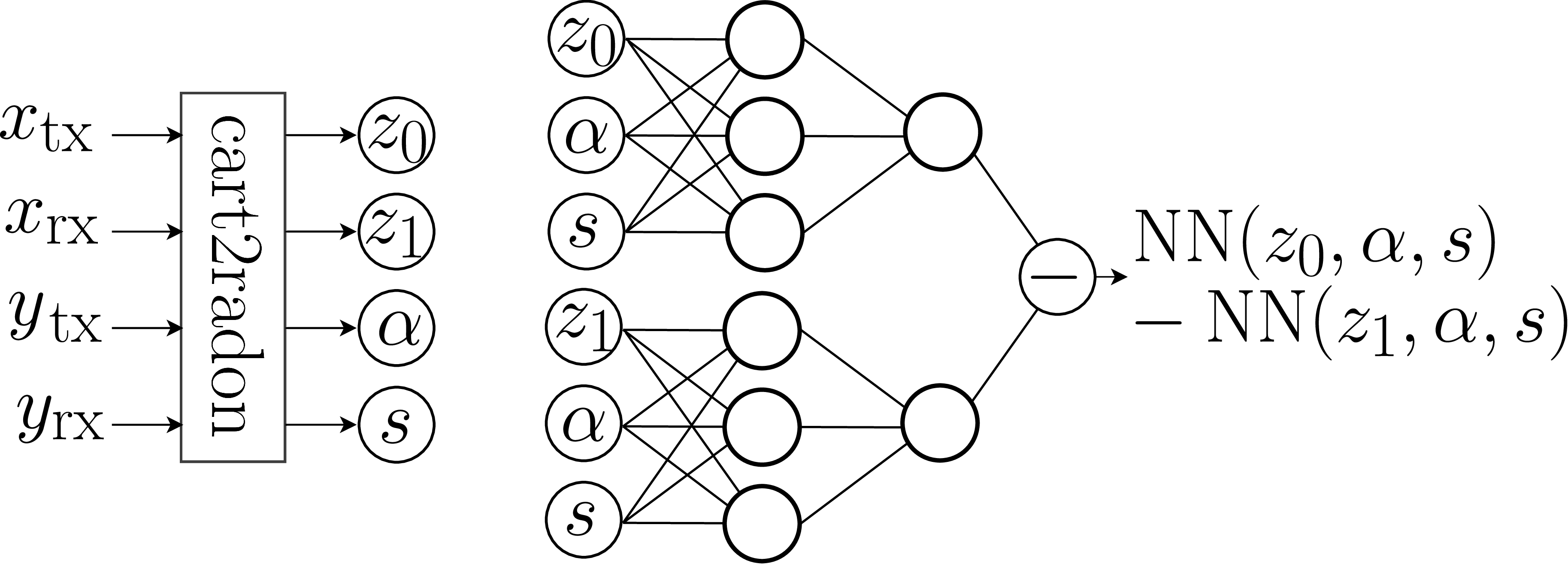}
}
\caption{Visualization of NN architecture for (a) the base MLP model with fourier features, (b) ISLF prediction using signed sum of NN evaluations.}
\label{fig:nn_architectures}
\end{figure}

The corresponding loss function in Radon-domain is given by
\begin{align}\label{equ:loss_radon}
\nonumber 
& \text{Loss}(\bg{\theta}) = \\ \nonumber 
& \sum_i \phi\Bigl( \text{SLF}(z^{(i)},\alpha^{(i)},s^{(i)}) - \frac{d}{dz} \text{NN}(\bg{\theta},z^{(i)},\alpha^{(i)},s^{(i)}) \Bigr) \\ \nonumber 
& + \sum_j \rho\Bigl( \text{ISLF}(z^{(j)}_{0},z^{(j)}_{1},\alpha^{(j)},s^{(j)}) \\ 
& - \bigl(\text{NN}(\bg{\theta},z^{(j)}_{0},\alpha^{(j)},s^{(j)}) - \text{NN}(\bg{\theta},z^{(j)}_{1},\alpha^{(j)},s^{(j)}) \bigr) \Bigr).
\end{align}
The transform of $(x,y)$-coordinates to Radon coordinates is given by
\begin{align}
& (z,\alpha,s) \to (x,y): \\ 
& (x,y) = (z \sin \alpha + s \cos \alpha, -z \cos \alpha + s \sin \alpha),
\end{align}
and the integration domain $[x_\text{tx},x_\text{rx}] \times [y_\text{tx},y_\text{rx}]$ is transformed according to the bijection $(x_\text{tx},y_\text{tx},x_\text{rx},y_\text{rx}) \to (z_0,z_1,\alpha,s)$:
\begin{align}\label{equ:radon_transf0}
\alpha = \text{atan2}(y_\text{rx} - y_\text{tx}, x_\text{rx} - x_\text{tx}) + \frac{\pi}{2}.
\end{align}
and
\begin{align}\label{equ:radon_transf}
\begin{bmatrix} z_0 \\ z_1 \\ s \end{bmatrix}
 =
\begin{bmatrix} 
\sin \alpha & 0 & \cos \alpha \\ 
-\cos \alpha & 0 & \sin \alpha \\
0 & \sin \alpha & \cos \alpha \\
0 & -\cos \alpha & \sin \alpha 
 \end{bmatrix}^{\dagger}
 \begin{bmatrix} x_\text{tx} \\ y_\text{tx} \\ x_\text{rx} \\ y_\text{rx} \end{bmatrix},
\end{align}
where $(\cdot)^\dagger$ denotes the Moore-Penrose pseudo inverse.

The resulting steps to obtain pathloss predictions using a trained pinn with optimized parameters $\bg{\theta}*$ are depicted in Fig. \ref{fig:nn_architectures} and summarized in Alg. \ref{alg:nn_inf} 

\begin{algorithm}[h]\label{alg:nn_inf}
\SetKwInput{KwData}{\textbf{Input}}
\SetKwInput{KwResult}{\textbf{Output}}
\KwData{Trained pinn with parameters $\bg{\theta}^*$, \\ tx-rx locations $\bm{x}_\text{tx}:=(x_\text{tx},y_\text{tx}),\bm{x}_\text{rx}:=(x_\text{rx},y_\text{rx})$}
\KwResult{$\text{SLF}(\bm{x})$, $\text{ISLF}(\bm{x}_{tx},\bm{x}_{rx})$}
\begin{enumerate}
\item Apply cartesian to Radon transform (\ref{equ:radon_transf0},~\ref{equ:radon_transf})
\begin{itemize}
\item map point $(x,y)$ to $(z,\alpha,s)$
\item map line segment $\overline{(x_\text{tx},y_\text{tx}),(x_\text{rx},y_\text{rx})}$ to $(z_0,z_1,\alpha,s)$
\end{itemize}

\item Predict $\text{SLF}: (z,\alpha,s) \mapsto \frac{d}{dz} \text{NN}(\bg{\theta}^\star,z,\alpha,s)$
\item Predict $\text{ISLF}: (z_0,z_1,\alpha,s) \mapsto \text{NN}(\bg{\theta}^\star,z_{0},\alpha,s) - \text{NN}(\bg{\theta}^\star,z_{1},\alpha,s)$
\end{enumerate}
\caption{Loss field and pathloss prediction using trained pinn.}
\end{algorithm}

\section{Simulation Results}
\label{sec:results}

\begin{figure}
	\includegraphics[width=0.95\columnwidth]{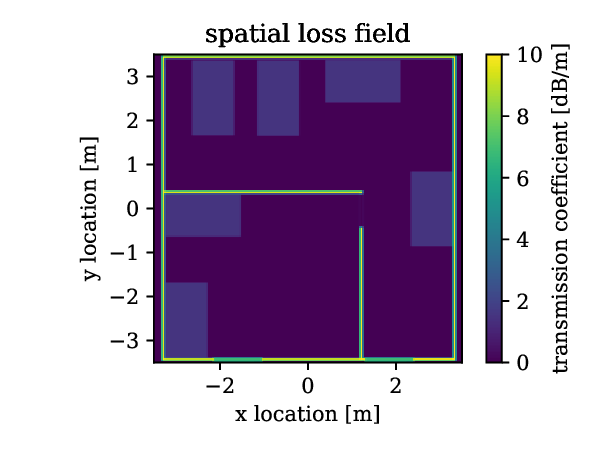}
	\caption{Spatial loss field used for training based on a building floor plan.}
	\label{fig:SLF1}
    \vspace{-1em}
\end{figure}

To benchmark the physics-informed training approach we model a spatial loss field according to an actual building floor plan with color encoding the approximate attenuation of different building materials as depicted in Fig. \ref{fig:SLF1}. The spatial loss field \ref{fig:SLF1} serves as a training input for the cost function \eqref{equ:loss_radon} and may be accompanied by pairwise pathloss measurements. In our simulation, pairwise pathloss measurements are excluded from the NN training but are used instead to estimate the transmit power and pathloss exponent in \eqref{equ:pl0}. Cost function, Radon-transform \eqref{equ:radon_transf}, NN forward and NN derivative operators are implemented in the JAX automatic differentiation system \cite{Bradbury2018} and the NN architecture is a 3-layer MLP with Fourier feature encoding, 256 neurons per layer, sigmoid activation function and normalized mean-squared error loss function. The training time of the final model was approximately \SI{1}{h} on an NVIDIA RTX A2000 and inference time per path of the trained model is approximately \SI{10}{\mu s}.

\begin{figure}
\centering \subfigure[proposed method]{
  \includegraphics[width=0.95\columnwidth]{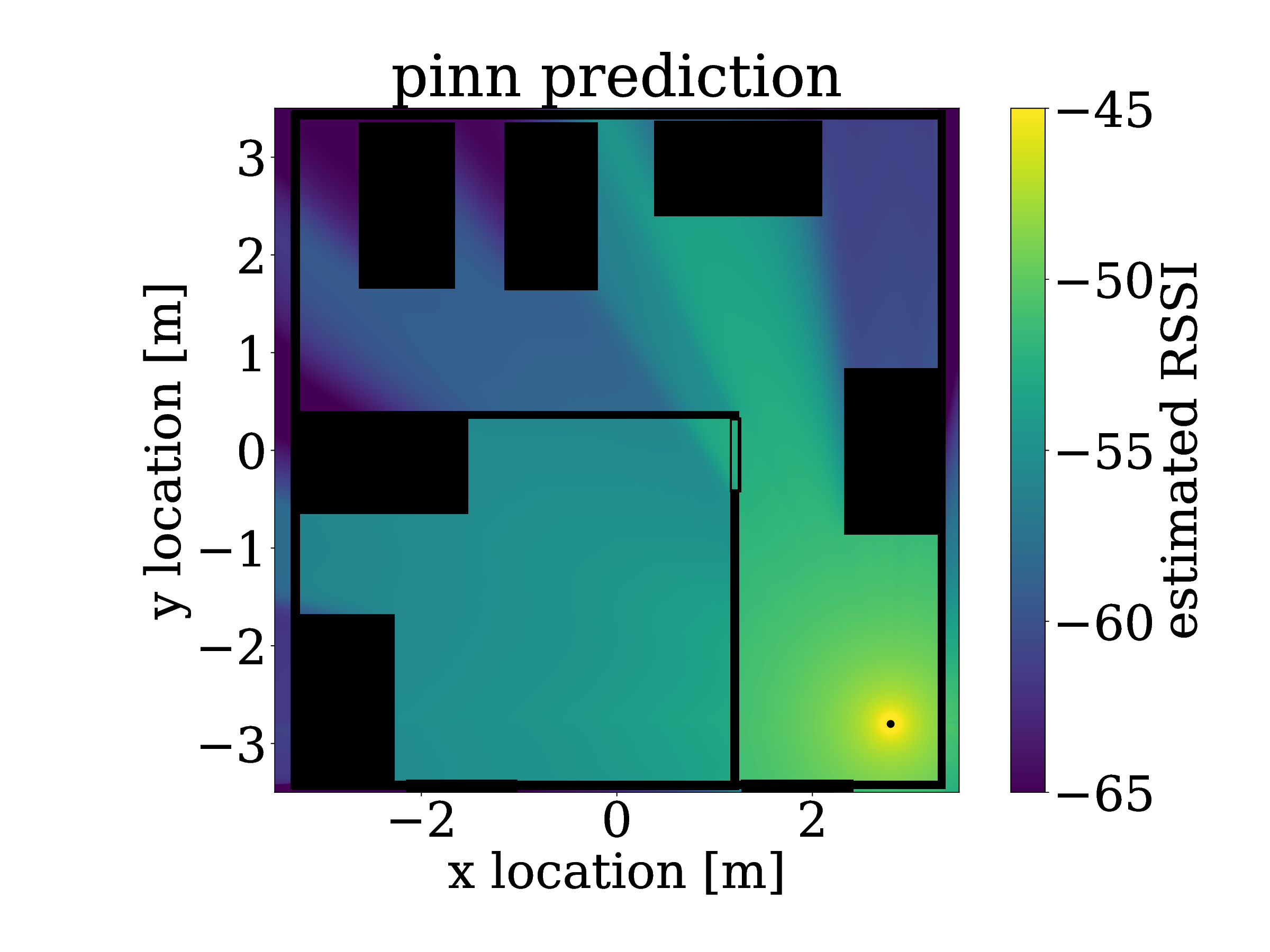}
}  \subfigure[Pylayers \cite{Amiot2013}]{
  \hspace{-1cm} \includegraphics[width=0.9\columnwidth]{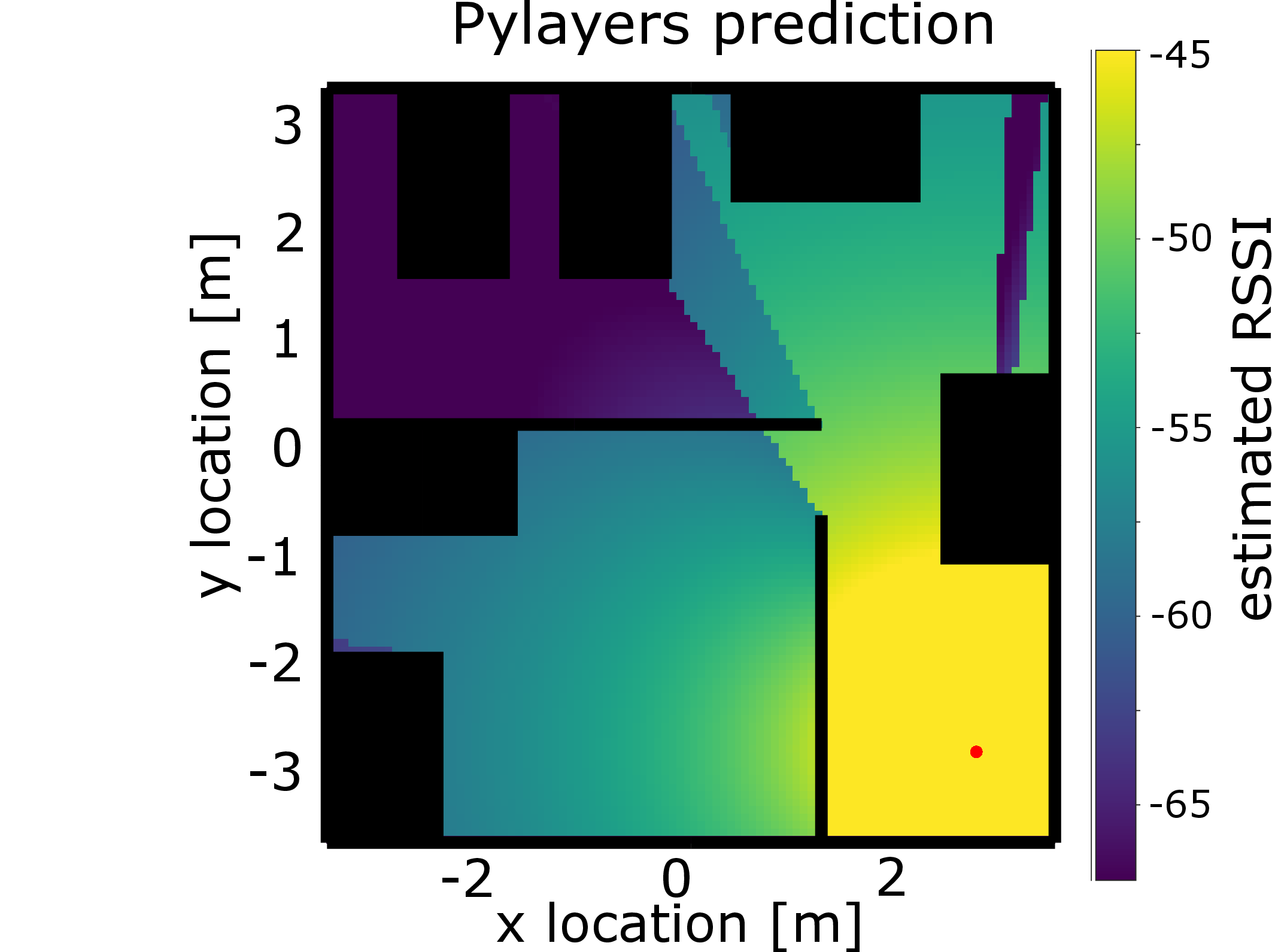}
}
\caption{Comparison of pathloss maps obtained by (a) proposed method and (b) Pylayers \cite{Amiot2013} with Motley-Keenan model.}
\label{fig:results_comparison}
\end{figure}

\begin{figure}
\centering \subfigure[]{
  \includegraphics[width=0.492\linewidth]{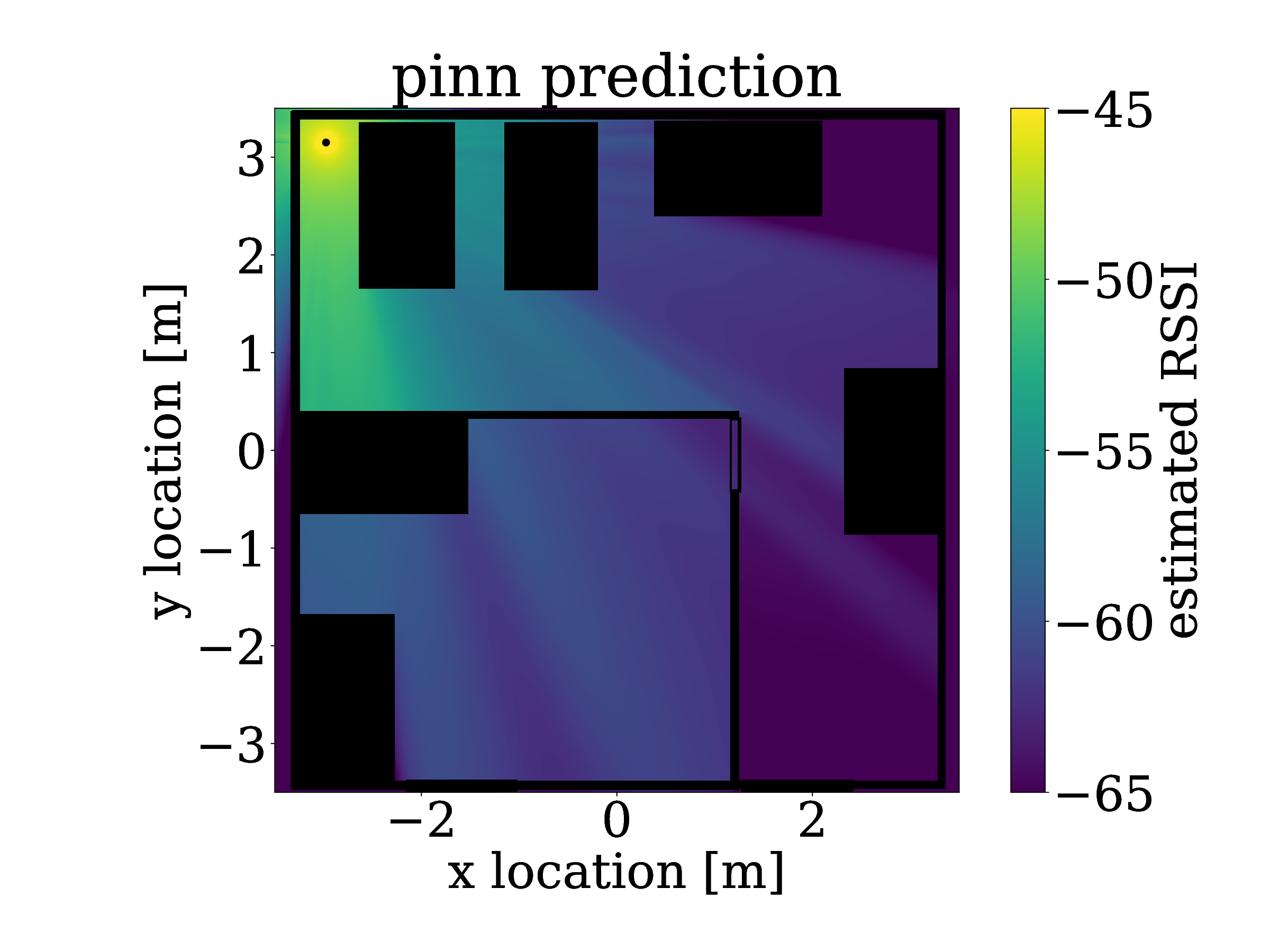}
} \hspace{-.55cm} \subfigure[]{
  \includegraphics[width=0.492\linewidth]{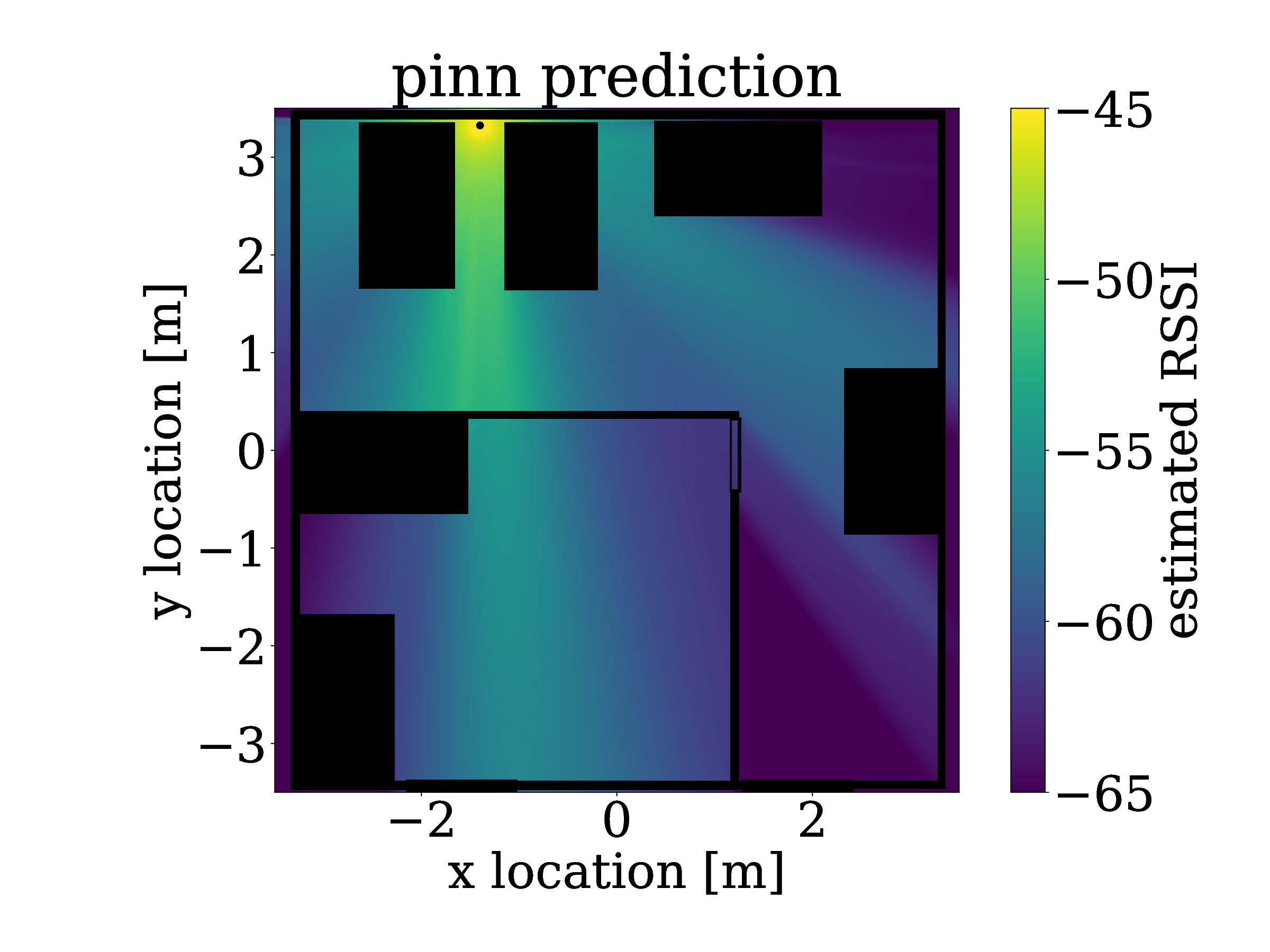}
}
\centering \subfigure[]{
  \includegraphics[width=0.492\linewidth]{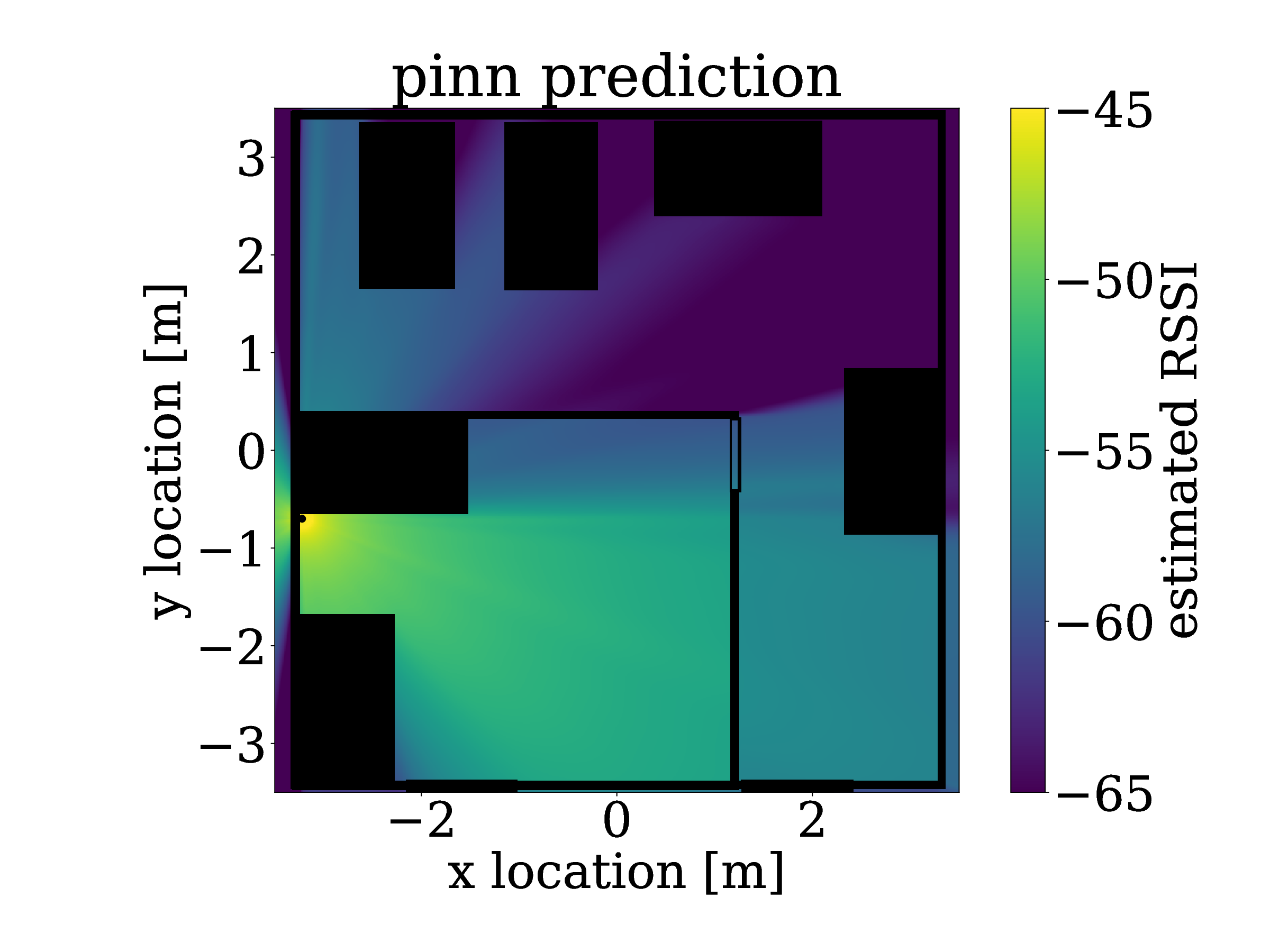}
} \hspace{-.55cm} \subfigure[]{
  \includegraphics[width=0.492\linewidth]{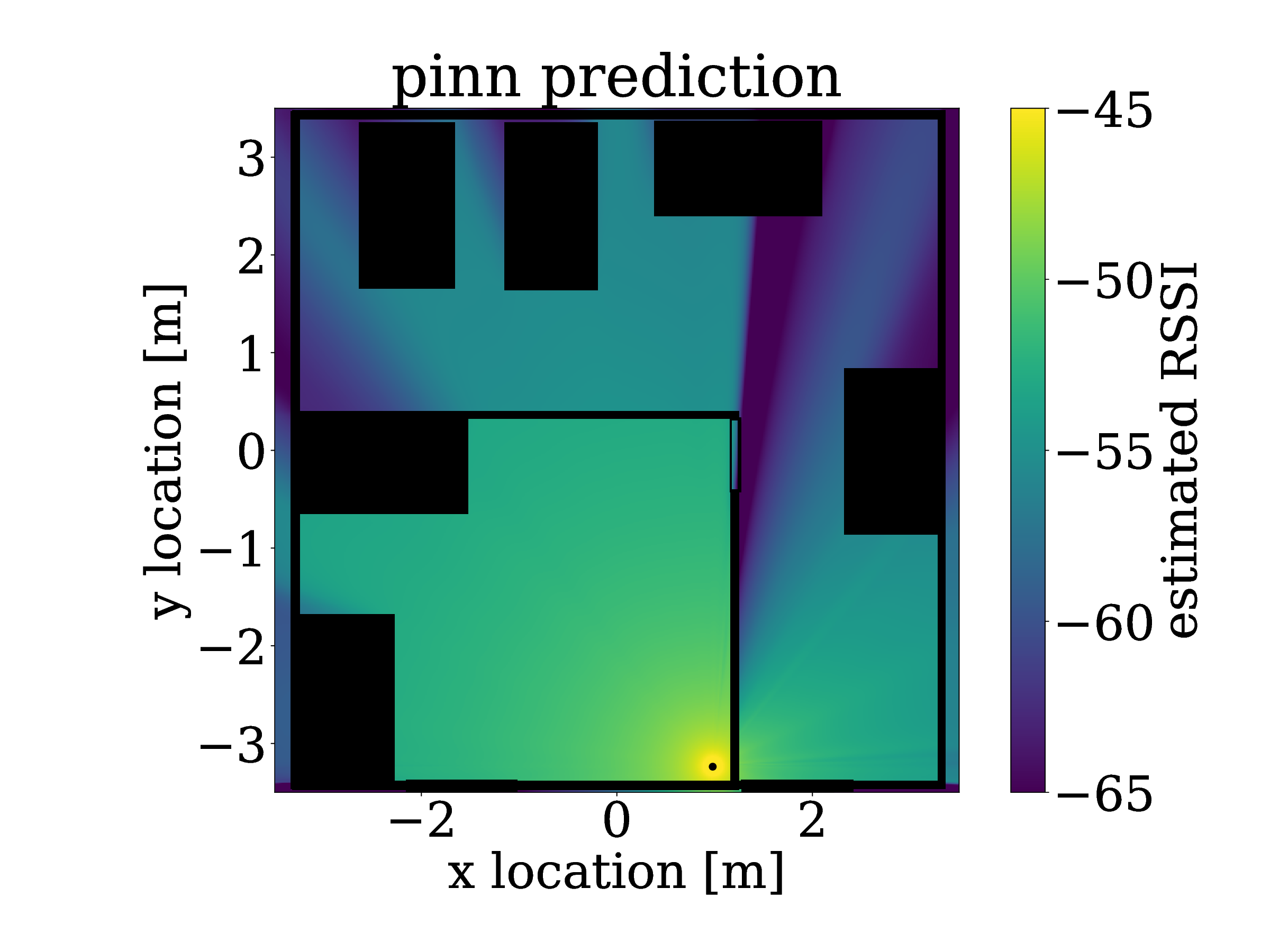}
}
\caption{Corner case test stage pathloss map using proposed method.}
\label{fig:results_cornercase}
\end{figure}

Fig. \ref{fig:results_comparison} depicts a comparison of the proposed pinn and a reference simulation using the Python Propagation and Localisation Simulator Pylayers \cite{Amiot2013} with a Motley-Keenan model. Additional test stage pathloss maps for corner cases using the proposed method are provided in Fig. \ref{fig:results_cornercase}.\footnote{An animated pathloss map for a prescribed transmitter trajectory is provided as supplementary material at \url{https://github.com/stli/2023_pinn/blob/main/pinn_prediction.gif}.}
It can be observed that the physics-informed neural network allows a fine-grained prediction including details of the propagation environment such as smaller attenuation by doors compared to drywalls and consistent estimation for different corner cases of transmitter locations

\section{Conclusion}
\label{sec:conclusion}
This paper introduced a new physics-aware machine learning method for pathloss prediction. While physics-aware learning methods have been mainly used for partial-differential equations (pdes), this paper introduced a novel formulation to solve an integral equation that appears in pathloss prediction and tomography applications. Compared to previous approaches, the proposed solution allows to include both information about the site geometry and recorded pathloss measurements between different transmitter and receiver pairs. By adhering to physical constraints, consistent pathloss predictions including fine-grained details of propagation environments can be obtained using standard neural networks.

\bibliographystyle{IEEEbib}
\bibliography{refs}

\begin{thebibliography}{10}

\bibitem{Anderson2004}
C.R. Anderson and T.S. Rappaport,
\newblock ``In-building wideband partition loss measurements at 2.5 and 60
  ghz,''
\newblock {\em IEEE Transactions on Wireless Communications}, vol. 3, pp.
  922--928, 2004.

\bibitem{Amiot2013}
N.~Amiot, M.~Laaraiedh, and B.~Uguen,
\newblock ``Pylayers: An open source dynamic simulator for indoor propagation
  and localization,''
\newblock in {\em 2013 IEEE International Conference on Communications
  Workshops (ICC)}. IEEE, 2013, pp. 84--88.

\bibitem{Levie2021}
R.~Levie, Ç. Yapar, G.~Kutyniok, and G.~Caire,
\newblock ``Radiounet: Fast radio map estimation with convolutional neural
  networks,''
\newblock {\em IEEE Transactions on Wireless Communications}, vol. 20, pp.
  4001--4015, 2021.

\bibitem{Patwari2008}
N.~Patwari and P.~Agrawal,
\newblock ``Nesh: A joint shadowing model for links in a multi-hop network,''
\newblock in {\em 2008 IEEE International Conference on Acoustics, Speech and
  Signal Processing}. IEEE, 2008, pp. 2873--2876.

\bibitem{Raissi2019}
M.~Raissi, P.~Perdikaris, and G.~Karniadakis,
\newblock ``Physics-informed neural networks: A deep learning framework for
  solving forward and inverse problems involving nonlinear partial differential
  equations,''
\newblock {\em Journal of Computational physics}, vol. 378, pp. 686--707, 2019.

\bibitem{Hennigh2021}
O.~Hennigh, S.~Narasimhan, M.~Nabian, A.~Subramaniam, K.~Tangsali, Z.~Fang,
  W.~Rietmann, M.and~Byeon, and S.~Choudhry,
\newblock ``Nvidia simnet™: An ai-accelerated multi-physics simulation
  framework,''
\newblock in {\em Computational Science--ICCS 2021: 21st International
  Conference, Krakow, Poland, June 16--18, 2021, Proceedings, Part V}.
  Springer, 2021, pp. 447--461.

\bibitem{Mueller2020}
T.~M{\"u}ller, F.~Rousselle, A.~Keller, and J.~Nov{\'a}k,
\newblock ``Neural control variates,''
\newblock {\em ACM Transactions on Graphics (TOG)}, vol. 39, no. 6, pp. 1--19,
  2020.

\bibitem{Lindell2021}
D.~Lindell, J.~Martel, and G.~Wetzstein,
\newblock ``Autoint: Automatic integration for fast neural volume rendering,''
\newblock in {\em Proceedings of the IEEE/CVF Conference on Computer Vision and
  Pattern Recognition}, 2021, pp. 14556--14565.

\bibitem{Hamilton2013}
B.~Hamilton, X.~Ma, R.~Baxley, and S.~Matechik,
\newblock ``Propagation modeling for radio frequency tomography in wireless
  networks,''
\newblock {\em IEEE Journal of Selected Topics in Signal Processing}, vol. 8,
  no. 1, pp. 55--65, 2013.

\bibitem{Maier2018}
A.~Maier, S.~Steidl, V.~Christlein, and J.~Hornegger,
\newblock ``Medical imaging systems: An introductory guide,''
\newblock 2018.

\bibitem{Bradbury2018}
J.~Bradbury, R.~Frostig, P.~Hawkins, M.~Johnson, C.~Leary, D.~Maclaurin,
  G.~Necula, A.~Paszke, J.~Vander{P}las, S.~Wanderman-{M}ilne, and Q.~Zhang,
\newblock ``{JAX}: composable transformations of {P}ython+{N}um{P}y programs,''
  2018.

\end{thebibliography}

\end{document}